\documentclass[letterpaper, 10 pt, conference]{ieeeconf}  

\IEEEoverridecommandlockouts                              

\usepackage[]{graphicx}
\usepackage{multirow}
\usepackage{blindtext}
\usepackage{subcaption}
\usepackage{xcolor}
\usepackage[export]{adjustbox}
\usepackage[normalem]{ulem}
\usepackage{float}
\usepackage{amssymb}
\usepackage{cite}
\usepackage{capt-of}
\usepackage{tabularx}

\title{\LARGE \bf
Stomach 3D Reconstruction Based on\\Virtual Chromoendoscopic Image Generation
}

\author{Aji Resindra Widya$^{1}$, Yusuke Monno$^{1}$, Masatoshi Okutomi$^{1}$,\\ Sho Suzuki$^{2}$, Takuji Gotoda$^{2}$, and Kenji Miki$^{3}$ 
\thanks{This work was supported by JSPS KAKENHI Grant Number 17H00744.}
\thanks{$^{1}$A. R. Widya, Y. Monno, and M. Okutomi are with the Department of Systems and Control Engineering, School of Engineering, Tokyo Institute of Technology, Meguro-ku, Tokyo 152-8550, Japan
(e-mail: aresindra@ok.sc.e.titech.ac.jp; ymonno@ok.sc.e.titech.ac.jp; mxo@sc.e.titech.ac.jp).}
\thanks{$^{2}$S. Suzuki and T. Gotoda are with the Division of Gastroenterology and Hepatology, Department of Medicine, Nihon University School of Medicine, Chiyoda-ku, Tokyo 101-8309, Japan.}
\thanks{$^{3}$K. Miki is with the Department of Internal Medicine, Tsujinaka Hospital Kashiwanoha, Kashiwa-city, Chiba 277-0871, Japan.}%
}

\begin{document}

\maketitle
\thispagestyle{empty}
\pagestyle{empty}

\begin{abstract}
Gastric endoscopy is a standard clinical process that enables medical practitioners to diagnose various lesions inside a patient's stomach. If any lesion is found, it is very important to perceive the location of the lesion relative to the global view of the stomach. Our previous research showed that this could be addressed by reconstructing the whole stomach shape from chromoendoscopic images using a structure-from-motion~(SfM) pipeline, in which indigo carmine~(IC) blue dye-sprayed images were used to increase feature matches for SfM by enhancing stomach surface's textures. However, spraying the IC dye to the whole stomach requires additional time, labor, and cost, which is not desirable for patients and practitioners. In this paper, we propose an alternative way to achieve whole stomach 3D reconstruction without the need of the IC dye by generating virtual IC-sprayed~(VIC) images based on image-to-image style translation trained on unpaired real no-IC and IC-sprayed images. We have specifically investigated the effect of input and output color channel selection for generating the VIC images and found that translating no-IC green-channel images to IC-sprayed red-channel images gives the best SfM reconstruction result.
\end{abstract}

\section{Introduction}
Gastric endoscopy is a well-applied clinical process that enables medical practitioners to find a gastric lesion, such as an ulcer and caner, from inside a patient's stomach.
Accurate localization of a found lesion is very important to decide the next clinical procedure. For example, if laparoscopic gastroectomy for early cancer needs to be done, the target cancer location relative to the global view of the stomach has to be known to decide the operative procedure. However, accurately recognizing the lesion's 3D location only from 2D endoscopic images is difficult for gastric surgeons, especially when the images are captured by another endoscopist.

In our previous study, we tackled the problem of lesion localization by reconstructing a whole stomach 3D shape from endoscopic images based on SfM~\cite{widya20193d, widya2019whole}. Although the stomach 3D reconstruction by SfM is very challenging because of texture-less stomach surfaces, we found that a whole stomach shape could be reconstructed by using red-channel images of chromoendoscopy with indigo carmine~(IC) blue dye, where the IC dye acts as an enhancement technique to bring up more textures to the stomach surface~\cite{alcantarilla2013enhanced}. Compared with other approaches such as barium radiography~\cite{yamamichi2016comparative} and 3D computed tomography gastrography~\cite{kim2015role}, our SfM-based approach could provide the stomach 3D model with both geometric and color texture information, which is a great advantage for accurate lesion localization. However, though the IC dye is commonly used in gastroendoscopy, spraying it on the whole stomach surface requires additional procedure time, labor, and cost, which is not desirable for both patients and practitioners. Furthermore, the IC dye may also hinder the visibility for some lesions and reconstructed stomach 3D models because of its dark color tone. 

In this paper, we propose a novel SfM-based approach for whole stomach 3D reconstruction that does not require chromoendoscopic image sequences. Instead of spraying the IC dye during endoscopy, we generate virtual IC-sprayed~(VIC) images from no-IC images based on image-to-image style translation with CycleGAN~\cite{zhu2017unpaired}. The SfM pipeline is then applied using the generated VIC images.

With the rise of deep learning, image-to-image translation, in which the goal is to learn the mapping between one style of images to another, is attracting attention from researchers. The style translation has been proven to be useful for medical applications such as in colonoscopy depth estimation~\cite{rau2019implicit}. It is also reported that generating VIC images using CycleGAN improves the lesion detection and classification performance in colonoscopy~\cite{fukuda2019generating}. Even though our VIC image generation is inspired by the study~\cite{fukuda2019generating}, we apply the generated VIC images for stomach 3D reconstruction, which is the first attempt to the best of our knowledge.

In our experiments, we trained CycleGAN for the style translation using no-IC single-color-channel images and IC-sprayed red-channel images and investigated the effect of input color channel selection. As a result, we found that the CycleGAN translating the no-IC green-channel images to the IC-sprayed red-channel images gives the best VIC images for SfM. Using those VIC images, we were able to reconstruct the whole stomach 3D model without the need of real IC-sprayed images. We can also localize an image frame including a gastric ulcer in the reconstructed 3D model.

\section{Materials and Methods}
\subsection{Endoscope video dataset} \label{sec:dataset}
In this work, we used exactly the same endoscope video dataset from our previous work~\cite{widya2019whole}. In the dataset, there are seven videos captured from seven subjects undergoing general gastroendoscopy procedure. Each video contains two different image type sequences: no-IC and IC-sprayed sequences. We extracted the image frames from each video and divided them based on its sequence type to obtain training image data for VIC generation and also test no-IC sequences for 3D reconstruction. The experimental protocol was approved by the research ethics committees of Tokyo Institute of Technology and Nihon University Hospital.

\subsection{Cycle-consistent image-to-image translation (CycleGAN)}\label{sec:cycleGAN}
Since the exact pair of no-IC and IC-sprayed images is impossible to obtain, we decided to use CycleGAN~\cite{zhu2017unpaired} as our image-to-image translator that supports unsupervised and unpaired training data. Let $A$ and $B$ be two different image domains. CycleGAN consists of two sets of generator and discriminator pair, $(G_A, D_A)$ and $(G_B, D_B)$. 
The generator's task is to generate a virtual image by translating an input image from one domain to another and fool its opposite domain's discriminator. 
In the other hand, the discriminator's task is to distinguish between the generated and real images. 

The loss of CycleGAN consists of discriminator, generator and cycle consistency loss. Let $b' = G_A(a)$ and $a' = G_B(b)$ describe the output of a generator from a random input image $a, b \sim p_{data}$, respectively. Each respective discriminator, $D_A$ and $D_B$, should give high scores for real input images, $a$ and $b$, and low scores for generated input images, $a'$ and $b'$, respectively. The consistency loss makes sure that CycleGAN generates a close image with the original input image when translating it circularly, e.g., $a \approx a_{cyc}$, where $a_{cyc} = G_B(G_A(a))$. The consistency loss enables CycleGAN to be trained on the unpaired set of images for image-to-image style translation.

\subsection{Generating VIC images using CycleGAN}\label{sec:generation}

\begin{figure}
    \centering
    \vspace{2mm}
    \includegraphics[width=0.98\columnwidth]{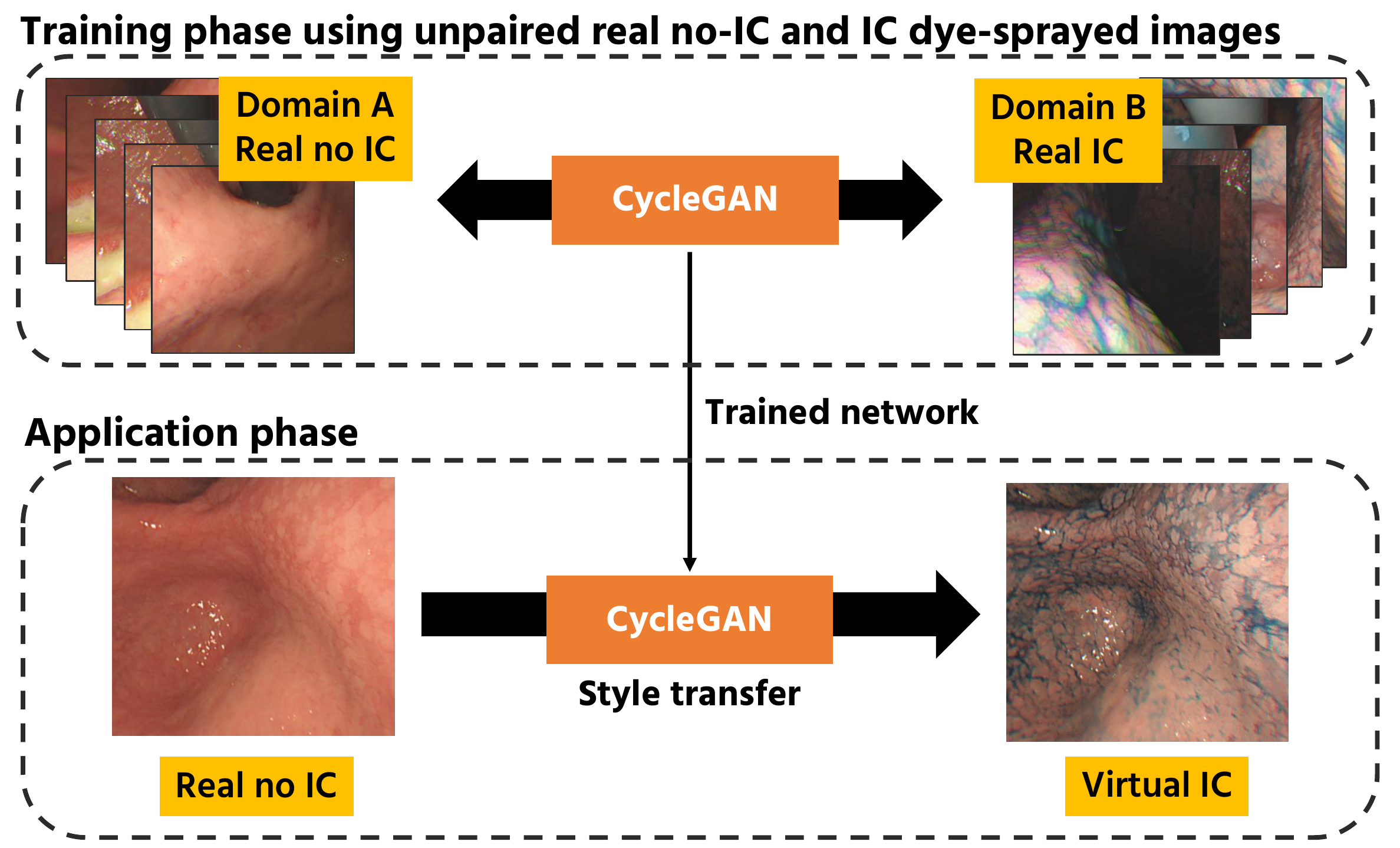}
    \caption{Our CycleGAN training and application overview}
    \label{fig:texturetransfer}
\end{figure}

Figure~\ref{fig:texturetransfer} shows our CycleGAN training and application overview~(the RGB color image case is shown for better visualization, though we actually use single-channel images as explained below). We train CycleGAN to translate the styles between no-IC and IC-sprayed images using the previously mentioned image dataset.

In our previous research~\cite{widya2019whole}, we observed that there is color channel misalignment in the RGB data. Because of that, we used single-channel images for the 3D reconstruction and investigated which color channel gives the best 3D reconstruction result. It was found that the whole stomach could be reconstructed using IC-sprayed red-channel images.
This is because the red channel of IC-sprayed images has the best contrast and the most visible textures. We also found that, for the case of no-IC images, the green channel gave the best 3D reconstruction result, though only partial stomach could be reconstructed. The blue channel was not preferable for 3D reconstruction due to low contrasts.

Based on those findings, we set the target domain to the IC-sprayed red-channel, since it is the best channel for stomach reconstruction. We then conducted two separated CycleGANs training using the no-IC red and green channels to investigate which color channel is better for the input to generate VIC images for SfM. Specifically, we set the domain $A$ and $B$ for each CycleGAN to (i)~no-IC red-channel and IC-sprayed red-channel and (ii)~no-IC green-channel and IC-sprayed red-channel. For the rest of this paper, we will refer to each CycleGAN as cGAN\textsubscript{r2r} and cGAN\textsubscript{g2r}, respectively. We then applied each trained CycleGAN to generate the VIC images from no-IC images.

\subsection{3D reconstruction using the generated VIC images}\label{sec:3dRec}
We followed the 3D reconstruction pipeline presented in our previous research~\cite{widya2019whole}. It consists of point cloud reconstruction, outlier removal, and mesh and texture generation. The point cloud reconstruction follows the general flow of SfM~\cite{schonberger_structure--motion_2016}. It starts with extracting scale invariant feature transform~(SIFT) features~\cite{lowe_distinctive_2004} from all of the input images followed by exhaustive feature matching. Those steps are then followed by features triangulation, poses estimation, and bundle adjustment~\cite{triggs1999bundle} in parallel. It is then followed by random sample consensus~(RANSAC)-based plane-fitting outlier removal to remove apparent outlier 3D points. Meshing and texturing is also performed to obtain the textured 3D mesh model.

\section{Results and Discussion}

\subsection{Implementation details}
We trained CycleGAN using a single NVIDIA GeForce GTX 1080Ti GPU. We used the same network setting as presented in~\cite{zhu2017unpaired}. The network was trained for 100 epochs for each channel setting, i.e., cGAN\textsubscript{r2r} and cGAN\textsubscript{g2r}, with 7978 no-IC images and 7453 IC-sprayed images. The training took approximately 28 hours to complete. Due to the GPU memory limitation, we resized the original $1155\times1003$ images to $600\times524$ pixels and trained the CycleGANs with randomly cropped image patches of $510\times510$ pixels. For the 3D reconstruction pipeline, we used exactly the same setup and implementation with our previous research~\cite{widya2019whole}.

\subsection{VIC image generation results}
\begin{figure}
\centering
\vspace{2mm}
\begin{adjustbox}{max width = \columnwidth}
\begin{subfigure}{0.49\columnwidth}
    \centering
    \includegraphics[width=0.95\columnwidth]{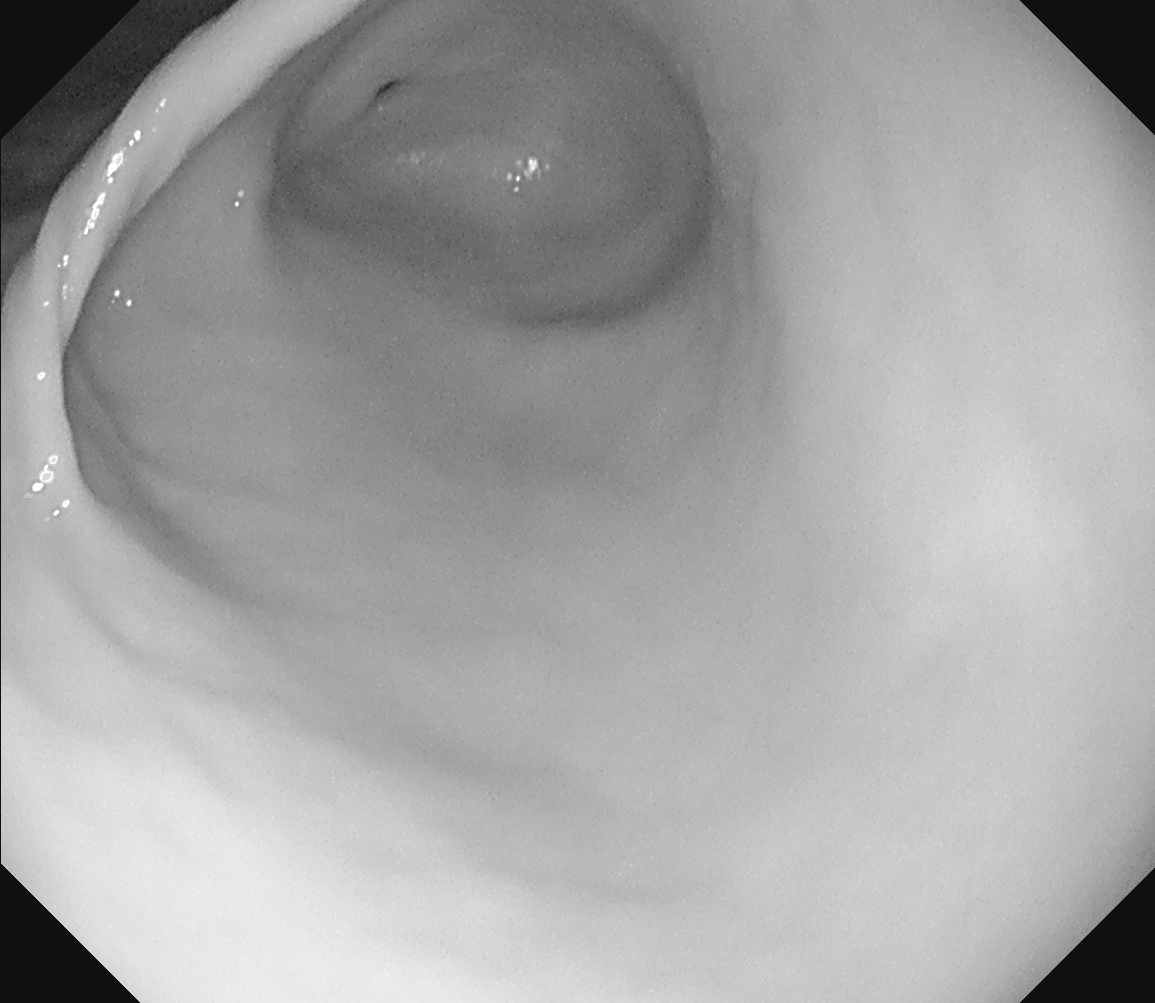} \\ \vspace{-1mm}
    \caption{No-IC red-channel}
\end{subfigure}
\begin{subfigure}{0.49\columnwidth}
    \centering
    \includegraphics[width=0.95\columnwidth]{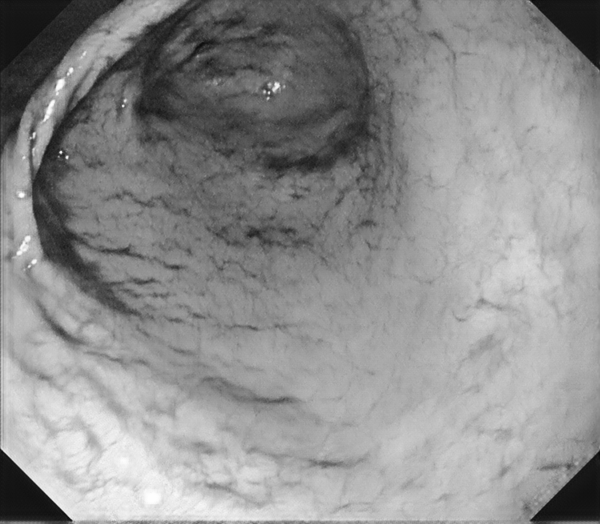} \\ \vspace{-1mm}
    \caption{VIC from (a) with cGAN\textsubscript{r2r}}
\end{subfigure}
\end{adjustbox}
\begin{adjustbox}{max width = \columnwidth}
\begin{subfigure}{0.49\columnwidth}
    \centering
     \vspace{2mm}
    \includegraphics[width=0.95\columnwidth]{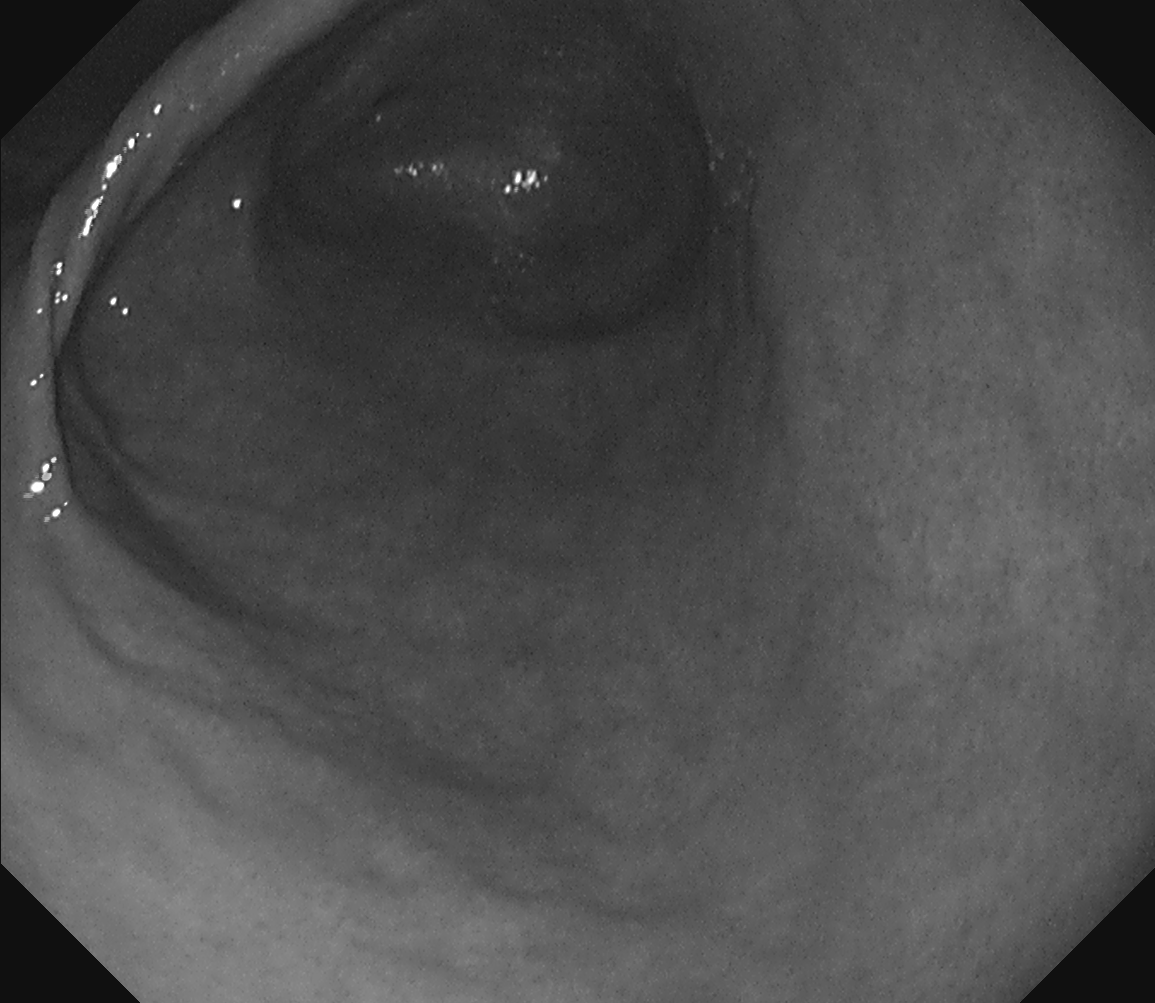} \\ \vspace{-1mm}
    \caption{No-IC green-channel}
\end{subfigure}
\begin{subfigure}{0.49\columnwidth}
    \centering
     \vspace{2mm}
    \includegraphics[width=0.95\columnwidth]{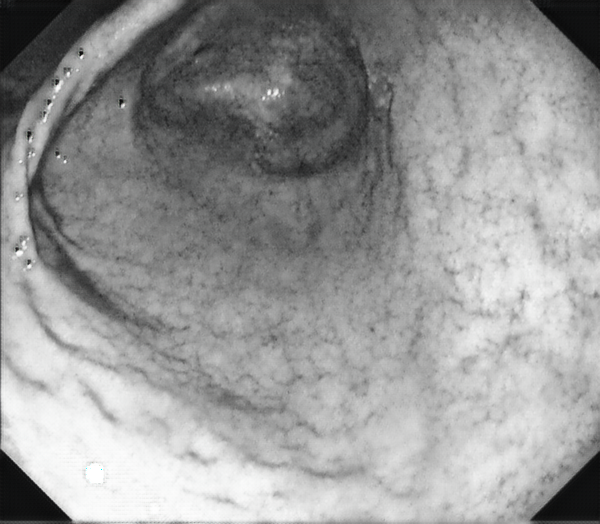} \\ \vspace{-1mm}
    \caption{VIC from~(c) with cGAN\textsubscript{g2r}}
\end{subfigure}
\end{adjustbox}
\vspace{-1mm}
\caption{Example results of generated VIC images.
}
\label{fig:comparisonimages}
\end{figure}

We first show the example results of generated VIC images using cGAN\textsubscript{g2r} and cGAN\textsubscript{r2r}. Figure~\ref{fig:comparisonimages} shows the comparison between the input no-IC images and the generated VIC images. The images~(a) and~(b) show the input no-IC red-channel image and the output VIC red-channel image by cGAN\textsubscript{r2r}, respectively. The images~(c) and~(d) show the input no-IC green-channel image and the output VIC red-channel image by cGAN\textsubscript{g2r}, respectively.

As we can see from the results, both cGAN\textsubscript{r2r} and cGAN\textsubscript{g2r} can transfer the style of the IC-sprayed image to the input no-IC image, not only by transferring the IC pattern, but also either tuning up or down the overall surface brightness. However, if we see the no-IC red-channel image of~(a), we can observe that the surface looks fairly texture-less. It is hard even for convolutional neural networks~(CNN) to extract features from this kind of texture-less images. In other hand, the no-IC green-channel image of (c) shows more textures, enabling slightly better style transfer. We will discuss the effect of the input channel in more detail in the following subsection.

\subsection{Feature matching results}
Using the generated VIC images for Subject~B, we calculated the average number of extracted SIFT features per image. The VIC images from cGAN\textsubscript{r2r} have the average of $3003.05$ features, while the VIC images from cGAN\textsubscript{g2r} have the average of $3511.49$ features. As the baselines, we also calculated the average numbers of extracted SIFT features of no-IC red-channel and green-channel images, which are $633.75$ and $743.18$ features, respectively. It is clear that the VIC images have more extracted features compared to the no-IC images by more than four times.

\begin{figure}[t!]
\centering
\vspace{2mm}
\includegraphics[width=0.9\columnwidth]{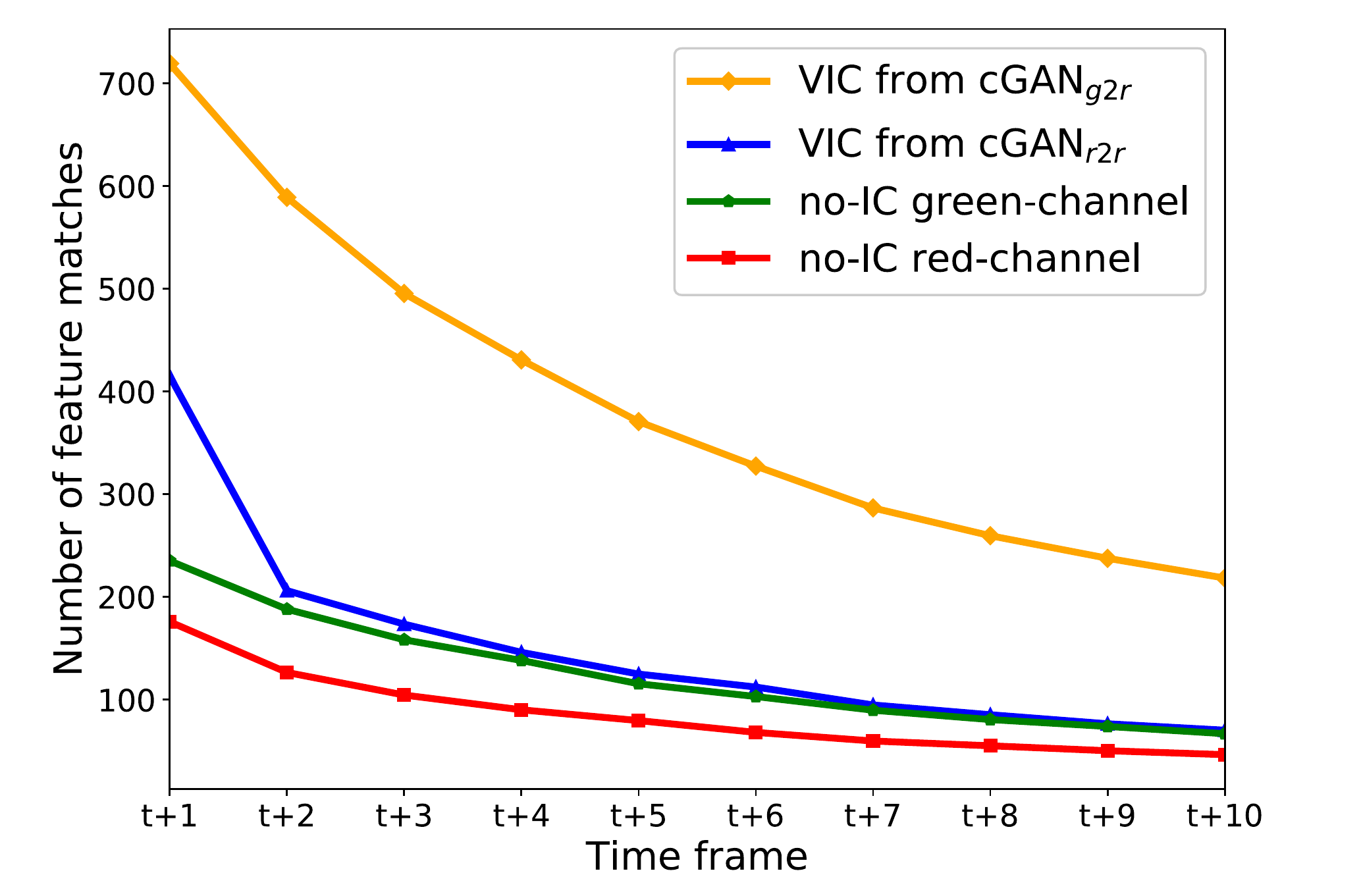}
\vspace{-2mm}
\caption{Comparison of the average number of feature matches between the anchor frame and its 10 consecutive frames. The x-axis represents the relative time stamp and the y-axis represents the average number of matches calculated using 43 samples. It is clearly shown that the VIC images from cGAN\textsubscript{g2r} has a higher number of matches across frames.}
\label{fig:featurematchesgraph}
\end{figure}

\begin{figure}[t!]
    \centering
    \begin{subfigure}{0.75\columnwidth}
        \centering
        \includegraphics[width=\columnwidth]{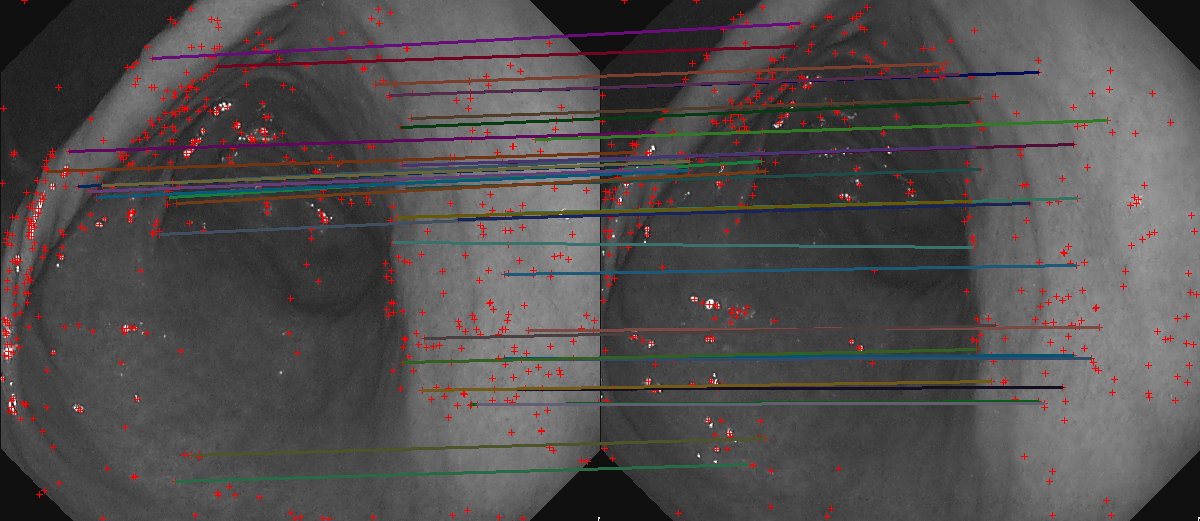} \\ 
        \vspace{-1mm}
        \caption{No-IC green-channel images}
    \end{subfigure}
    \\
    \begin{subfigure}{0.75\columnwidth}
        \centering
        \vspace{2mm}
        \includegraphics[width=\columnwidth]{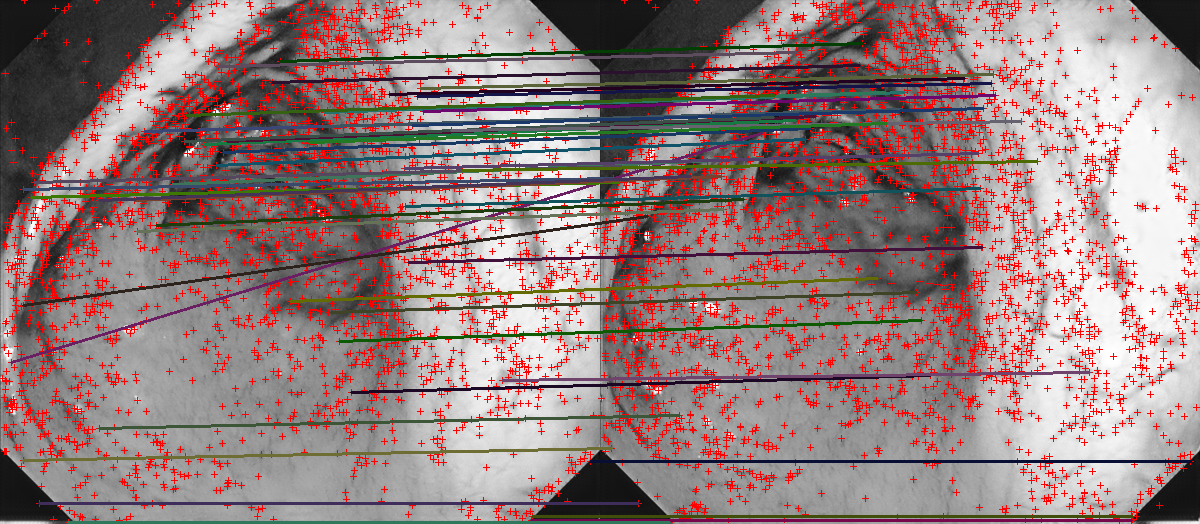} \\ 
        \vspace{-1mm}
        \caption{VIC images from cGAN\textsubscript{r2r}}
    \end{subfigure}
    \\
    \begin{subfigure}{0.75\columnwidth}
        \centering
        \vspace{2mm}
        \includegraphics[width=\columnwidth]{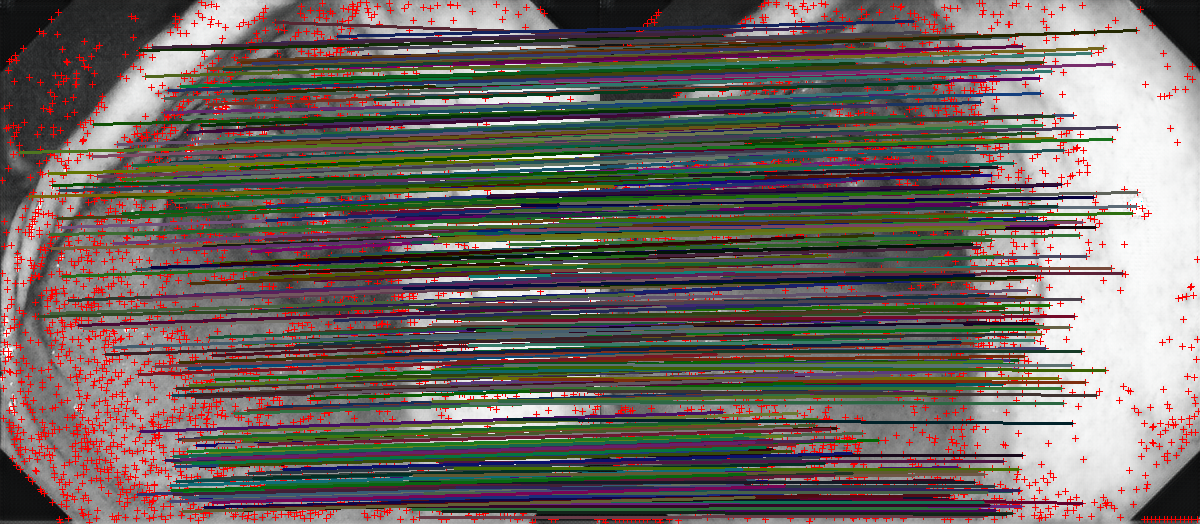} \\ 
        \vspace{-1mm}
        \caption{VIC images from cGAN\textsubscript{g2r}}
    \end{subfigure}
    \caption{The example feature matching result for two frames ($t$~and~$t+9$). The red marks represent the locations of extracted SIFT features. The color lines represent the matched features. It is clear that the number of feature matches in~(b) is much fewer than that in~(c), even though the number of extracted features in~(b) significantly increases  from~(a). This result implies that the generated VIC images from cGAN\textsubscript{g2r} have better pattern consistency between the frames.}
    \vspace{-4mm}
    \label{fig:featurematches}
\end{figure}

\begin{figure*}[t!]
\centering
\vspace{2mm}
    \begin{subfigure}{0.32\textwidth}
        \centering
        \includegraphics[width=0.8\columnwidth]{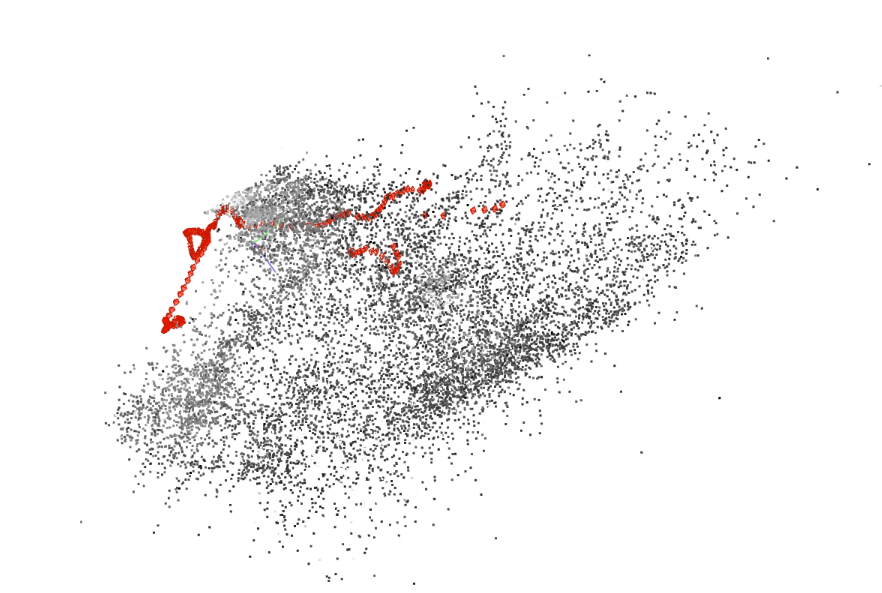} \\ \vspace{-2.2mm}
        \caption{No-IC green-channel images}
    \end{subfigure}
    \begin{subfigure}{0.32\textwidth}
        \centering
        \includegraphics[width=0.8\columnwidth]{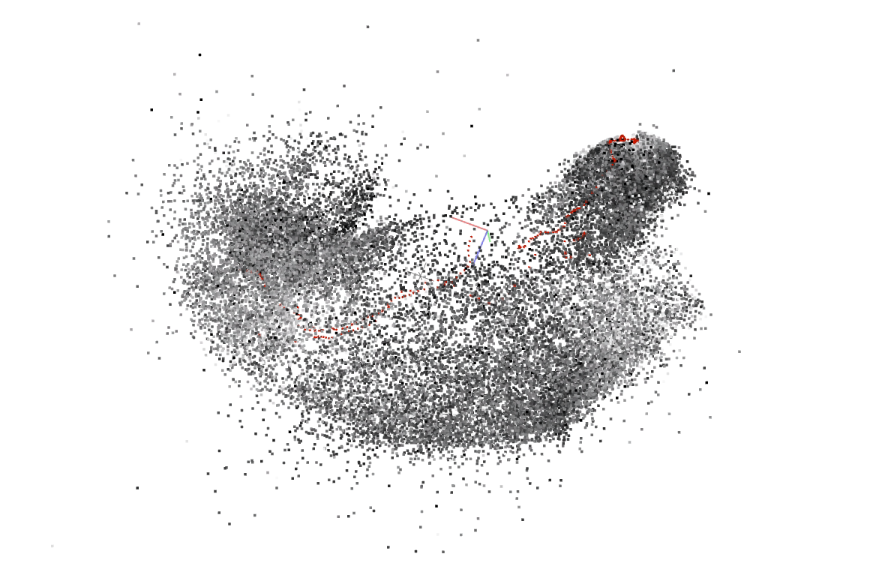} \\ \vspace{-1.5mm}
        \caption{VIC images from cGAN\textsubscript{r2r}}
    \end{subfigure}
    \begin{subfigure}{0.32\textwidth}
        \centering
        \includegraphics[width=0.8\columnwidth]{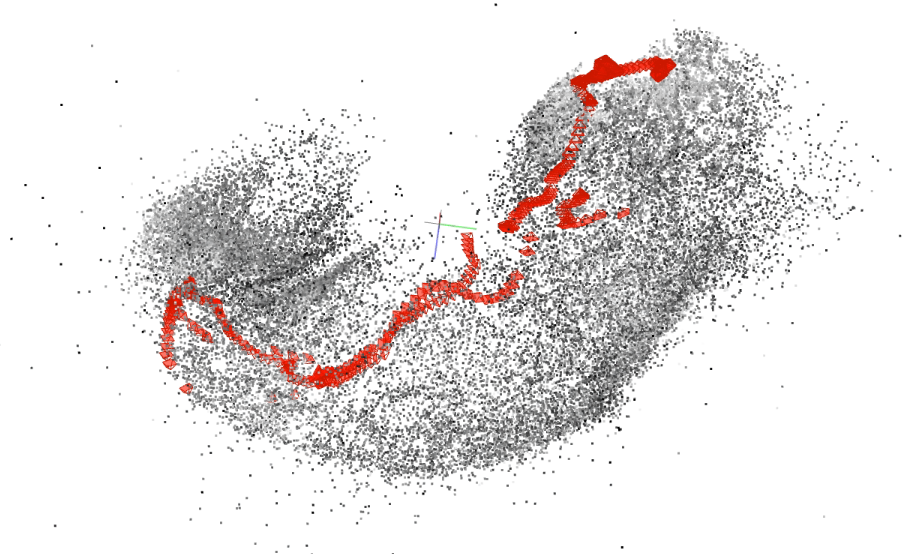} \\ \vspace{0.85mm}
        \caption{VIC images from cGAN\textsubscript{g2r}}
    \end{subfigure}\\ \vspace{1mm}
    \textrm{Point cloud reconstruction results for Subject B}\\
    
    \begin{subfigure}{0.32\textwidth}
        \centering
        \includegraphics[width=0.8\columnwidth]{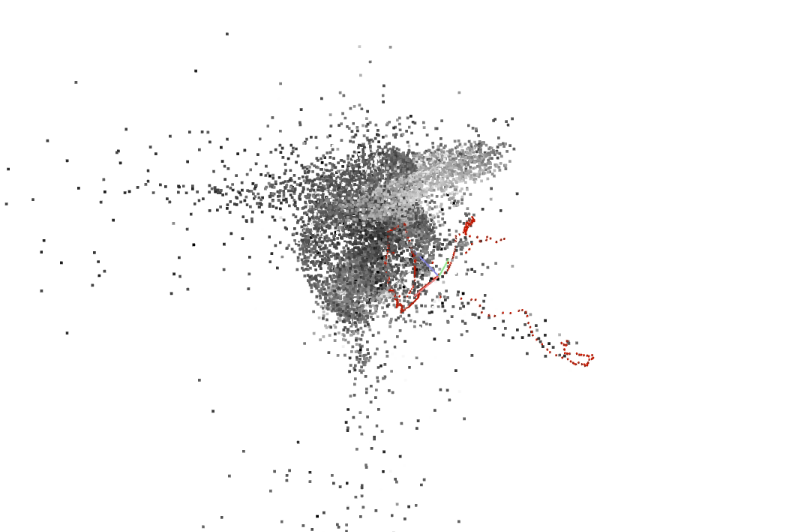} \\
        \caption{No-IC green-channel images}
    \end{subfigure}
    \begin{subfigure}{0.32\textwidth}
        \centering
        \includegraphics[width=0.8\columnwidth]{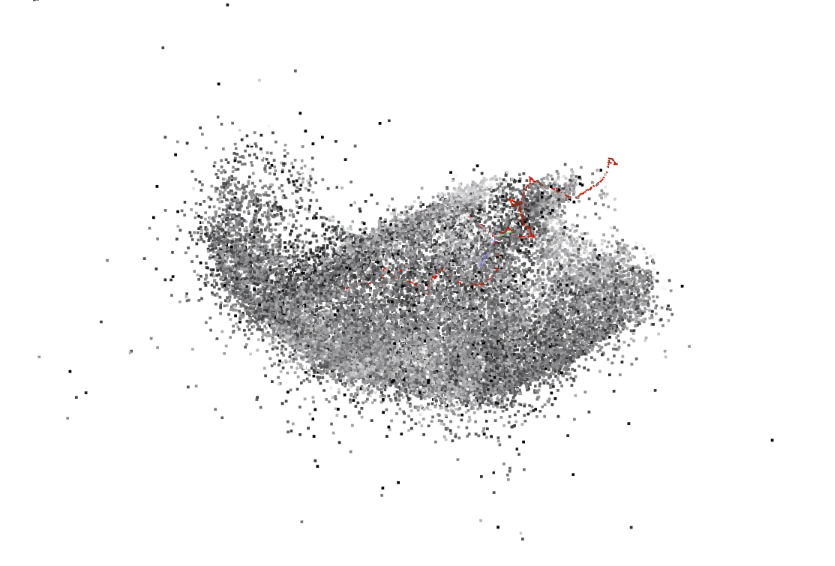} \\ \vspace{-1mm}
        \caption{VIC images from cGAN\textsubscript{r2r}}
    \end{subfigure}
    \begin{subfigure}{0.32\textwidth}
        \centering
        \includegraphics[width=0.8\columnwidth]{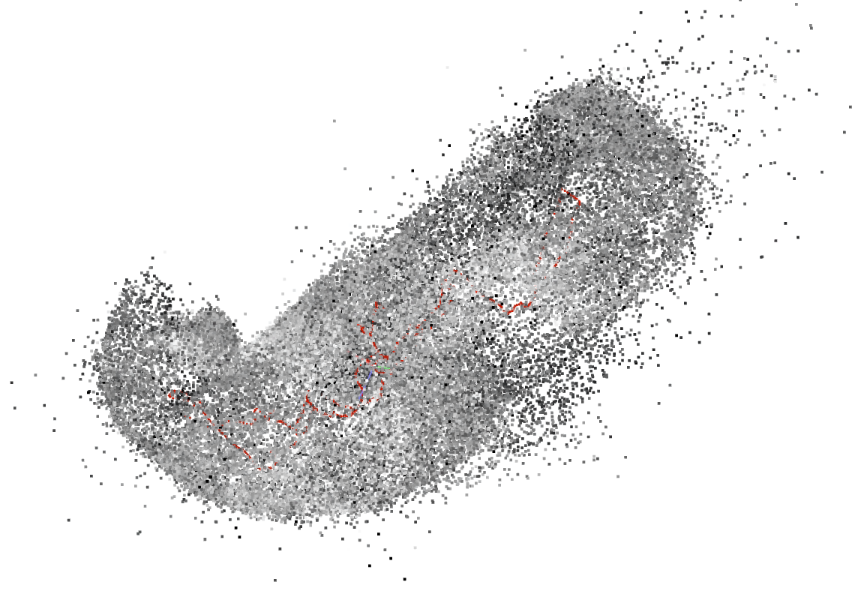} \\ \vspace{-1.6mm}
        \caption{VIC images from cGAN\textsubscript{g2r}}
    \end{subfigure}\\ \vspace{1mm}
    \textrm{Point cloud reconstruction results for Subject D}\\
    \caption{The SfM reconstruction results of Subject~B~(top) and Subject~D~(bottom) using no-IC green-channel images~(left), VIC images from cGAN\textsubscript{r2r}~(middle), and VIC images from cGAN\textsubscript{g2r}~(right). The gray dots represent the reconstructed 3D points and the red pyramids represent the estimated camera poses. Significant improvements from the baseline results of~(a) and~(b) are shown by the results of~(c) and~(f) using VIC images from cGAN\textsubscript{g2r}.}
    \label{fig:reconresult}
\end{figure*}

\begin{table*}[t!]
\centering
\vspace{2mm}
\caption{The objective evaluation of SfM results using no-IC green-channel images and VIC images from cGAN\textsubscript{g2r}.}
\label{tab:objectiveeval}
\begin{adjustbox}{max width=\textwidth}
\renewcommand{\arraystretch}{1.5}
\begin{tabular}{c|cccccccc}
\hline
\multicolumn{1}{l|}{} &
   &
  Subject A &
  Subject B &
  Subject C &
  Subject D &
  Subject E &
  Subject F &
  Subject G \\ 
  &
  Input images &
  2302 &
  439 &
  1715 &
  829 &
  1726 &
  1901 &
  1297 \\
  \hline
\multirow{3}{*}{\shortstack{No-IC\\green-channel\\images}}
  &
  Reconstructed images &
  2165 (94.05\%) &
  217 (49.43\%) &
  54 (3.15\%) &
  288 (34.74\%) &
  497 (28.79\%) &
  47 (2.47\%) &
  1248 (96.22\%) \\
 &
  3D points &
  204378 &
  13568 &
  2410 &
  14678 &
  27617 &
  2759 &
  105691 \\
 &
  Avg. observations &
  684.178 &
  524.668 &
  260.352 &
  353.312 &
  368.698 &
  339.766 &
  650.974 \\ \hline
\multirow{3}{*}{\shortstack{VIC images\\from cGAN\textsubscript{g2r}}}
 &
  Reconstructed images &
  \textbf{2201 (95.61\%)} &
  \textbf{438 (99.77\%)} &
  \textbf{1662 (96.91\%)} &
  \textbf{823 (99.28\%)} &
  \textbf{1668 (96.64\%)} &
  \textbf{1838 (96.69\%)} &
  \textbf{1297 (100\%)} \\
 &
  3D points &
  \textbf{412089} &
  \textbf{44866} &
  \textbf{207795} &
  \textbf{100115} &
  \textbf{238285} &
  \textbf{231744} &
  \textbf{213249} \\
 &
  Avg. observations &
  \textbf{2127.81} &
  \textbf{1007.01} &
  \textbf{1111.1} &
  \textbf{1099.95} &
  \textbf{1188.62} &
  \textbf{881.699} &
  \textbf{1504.79} \\ \hline
\end{tabular}
\end{adjustbox}
\end{table*}

However, solely increasing the number of features is not sufficient. Since SfM relies on the consistency of features across multiple images, we also tested the feature matching performance of the generated VIC images. For this purpose, we took 11 consecutive images from a sequence. We then chose the first image as an anchor, $t$, and performed feature matching to all of its consecutive images, $t+1, t+2, ..., t+10$.

Figure~\ref{fig:featurematchesgraph} shows the average number of feature matches between the anchor frame and each of its consecutive frames taken from 43 group-of-11-consecutive-images samples, which were extracted from the sequence of Subject~B.
It can be seen that the VIC images from cGAN\textsubscript{g2r} has a higher number of matches across frames compared to the other three image types. It implicitly means that the VIC images from cGAN\textsubscript{g2r} has better temporal pattern consistency between frames. Figure~\ref{fig:featurematches} shows the example feature matching results. 

\subsection{3D reconstruction results}

Figure~\ref{fig:reconresult} shows the SfM reconstruction results for Subject~B and~D using three different image types, i.e., no-IC green-channel images, VIC images from cGAN\textsubscript{r2r}, and VIC images from cGAN\textsubscript{g2r}. Please note that all three types of images were extracted or generated from the same source RGB sequence and thus the comparison can be fairly performed. Using those types of images, $49.43\%$, $88.84\%,$ and $99.77\%$ images of Subject B and $34.74\%$, $35.94\%,$ and $99.27\%$ images of Subject D were reconstructed, respectively. In~(a) and~(d), the stomach shape cannot be reconstructed using no-IC green-channel images. In~(b) and~(e), the results using VIC images from cGAN\textsubscript{r2r} only show partially reconstructed stomach shapes. In~(c) and~(f), we can confirm that the results using VIC images from cGAN\textsubscript{g2r} achieve the best point cloud quality and completeness. 

\newcommand{\rulesep}{\unskip\ \vrule\ }
\begin{figure*}[t!]
    \centering
    \vspace{2mm}
    \begin{subfigure}{0.65\textwidth}
    \begin{subfigure}{0.49\columnwidth}
        \centering
        \includegraphics[width=0.49\columnwidth]{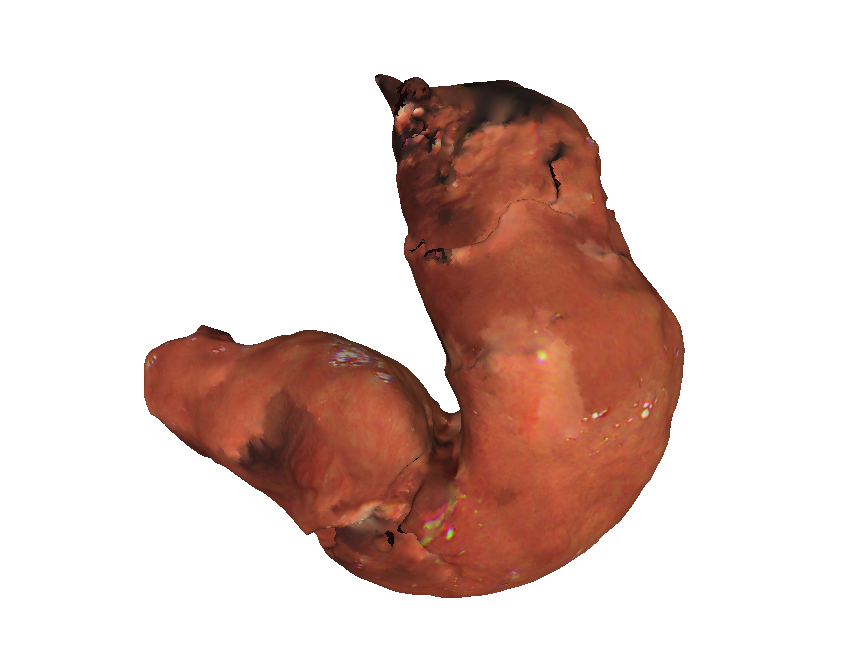}
        \includegraphics[width=0.49\columnwidth]{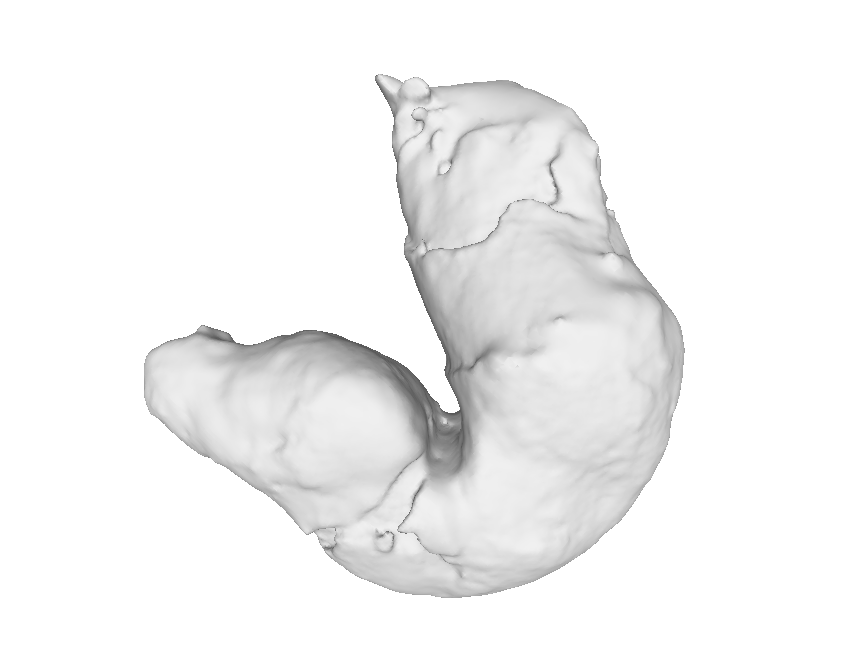}
        \caption{Subject A}
    \end{subfigure}
    \begin{subfigure}{0.49\textwidth}
        \centering
        \includegraphics[width=0.49\columnwidth]{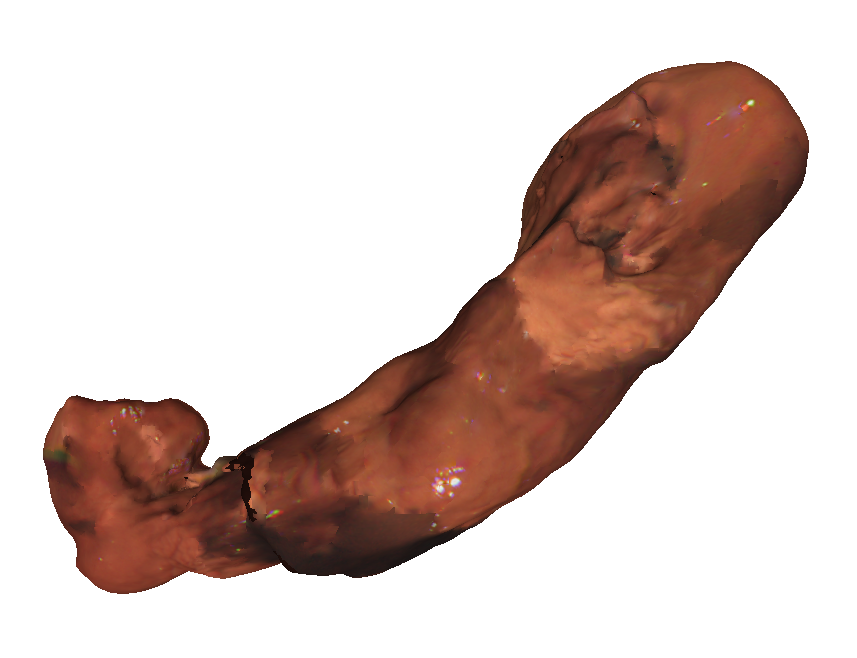}
        \includegraphics[width=0.49\columnwidth]{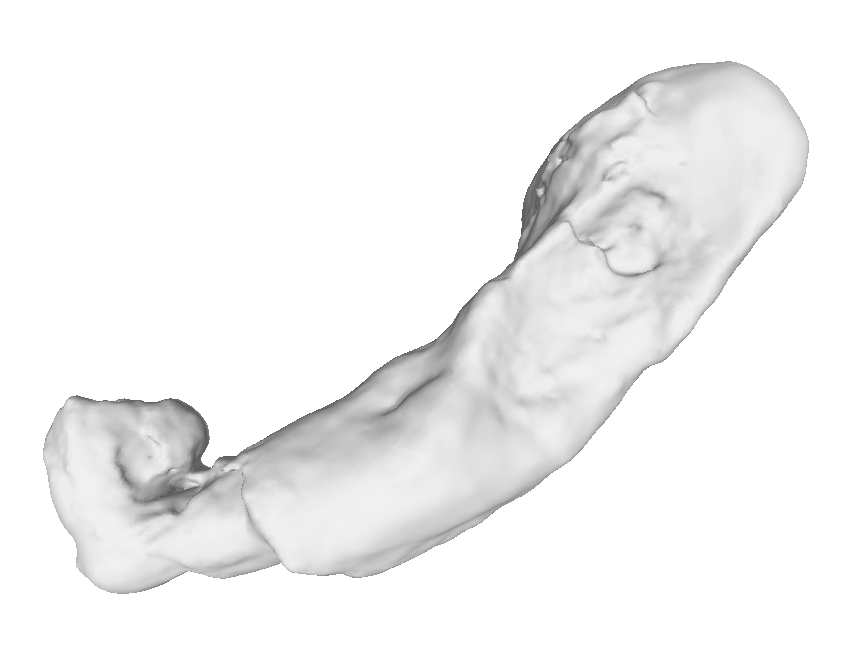}
        \caption{Subject C}
    \end{subfigure}
    \begin{subfigure}{0.49\textwidth}
        \centering
        \includegraphics[width=0.49\columnwidth]{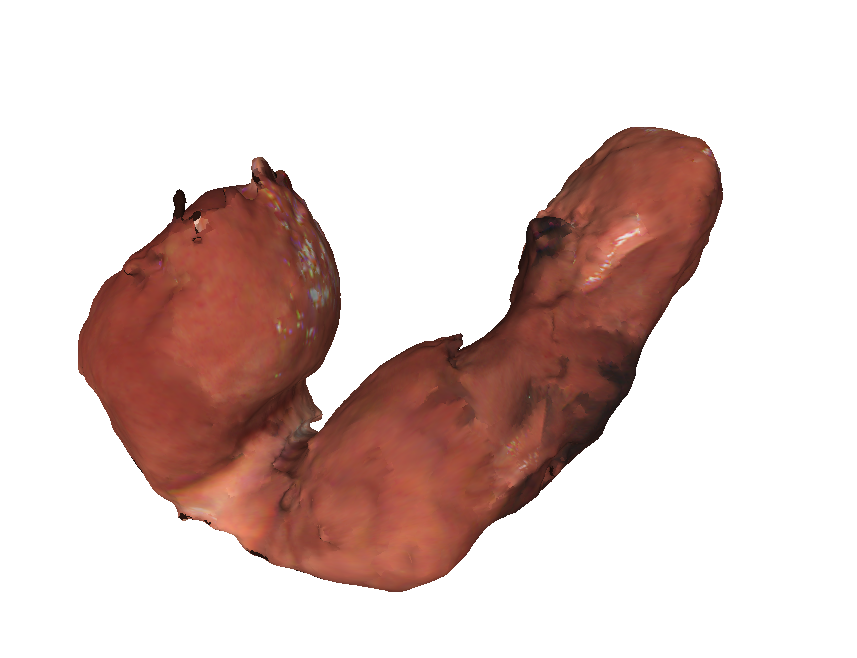}
        \includegraphics[width=0.49\columnwidth]{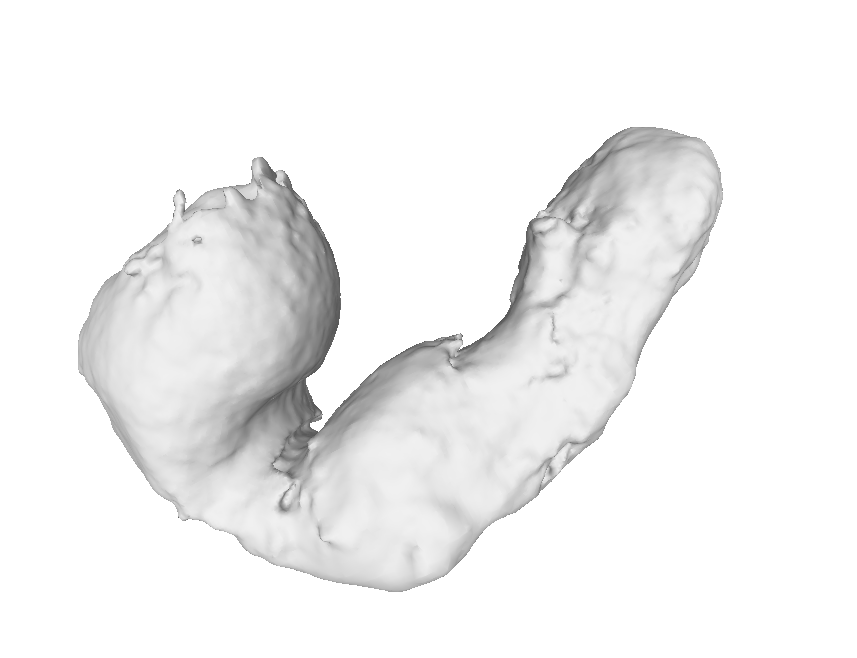}
        \caption{Subject E}
    \end{subfigure}
    \begin{subfigure}{0.49\textwidth}
        \centering
        \includegraphics[width=0.49\columnwidth]{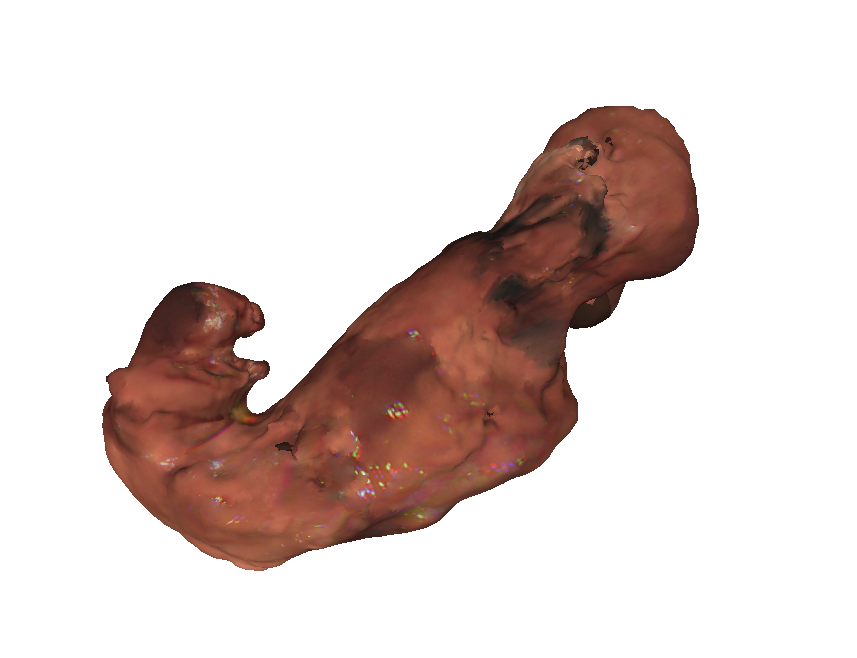}
        \includegraphics[width=0.49\columnwidth]{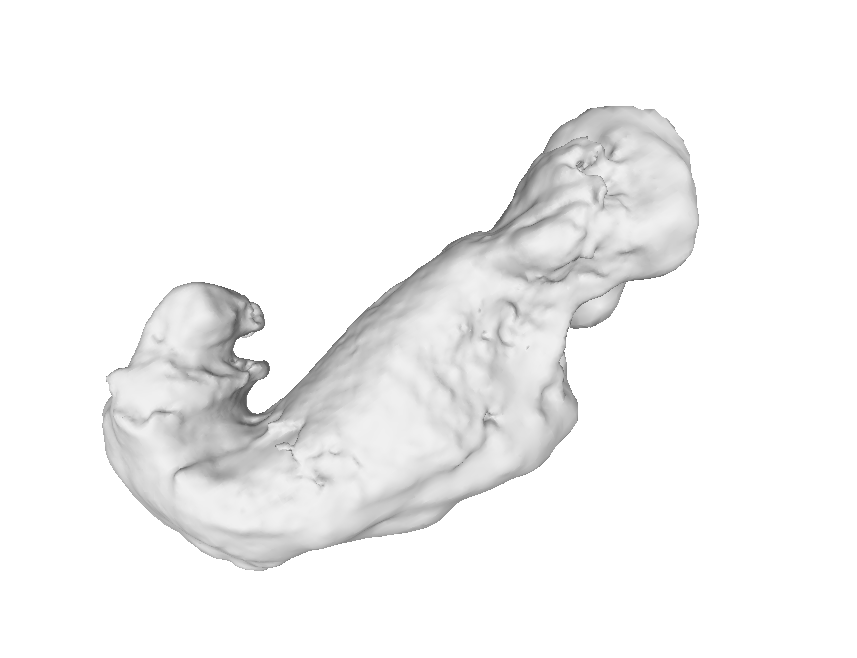}
        \caption{Subject F}
    \end{subfigure}
    \end{subfigure}
    \rulesep
    \begin{subfigure}{0.32\textwidth}
        \centering
        \includegraphics[width=0.8\columnwidth]{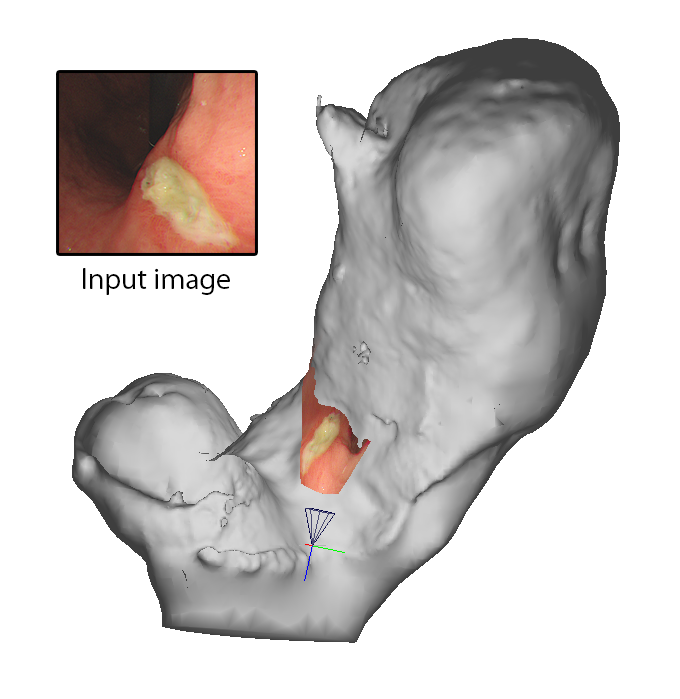}
        \vspace{1mm}
        \caption{Ulcer localization for Subject G}
        \label{fig:localization}
    \end{subfigure}
    \caption{The images (a) to~(d) show the meshing and texturing results using VIC images from cGAN\textsubscript{g2r}. We can confirm that all the results resemble the stomach shape, though the result of Subject~F is not as smooth as the others due to rigorous stomach movements. The image~(e) shows the lesion localization result for Subject G. We chose a reconstructed image containing the gastric ulcer to be localized. The estimated camera pose is shown as a blue pyramid. The corresponding RGB image of the localized camera was projected to the reconstructed mesh.}    
    \label{fig:meshing}
\end{figure*}

Table~\ref{tab:objectiveeval} shows the objective evaluation of SfM reconstruction results on all seven subjects. It shows that the generated VIC images from cGAN\textsubscript{g2r} achieve better results on all subjects compared to the baseline no-IC green-channel images. Using the VIC images for SfM significantly improves the number of reconstructed images, especially for Subject~B to~F. All reconstruction results using the VIC images achieve more than $95\%$ of reconstructed images. The triangulated 3D points also demonstrate significant improvement. It is because an increased number of feature matches across multiple images could be obtained from the VIC images. This led to the increase of features that could be triangulated, as shown by ``Avg. observation" in the table, which indicates the average number of triangulated feature points per image. 

Figure~\ref{fig:meshing}(a)--\ref{fig:meshing}(d) show the meshing and texturing results using the VIC images from cGAN\textsubscript{g2r}. We can see that the resulting meshes are well reconstructed and well textured. We can also confirm that the resulting meshes resemble the shape of the stomach. In Fig.~\ref{fig:meshing}(e), we also show the localization result of the gastric ulcer in Subject G. To localize the ulcer, we chose a reconstructed VIC image containing the ulcer and then projected the corresponding RGB image onto the reconstructed mesh using the estimated camera pose. We believe that our SfM-based approach could become a useful tool for the lesion localization.

\section{Conclusion}
In this paper, we have presented a new approach to reconstruct a whole stomach 3D shape from a gastroendoscopy video without the need of IC dye spraying. We have applied CycleGAN as an image-to-image style translator to generate VIC images from no-IC images for the stomach 3D reconstruction and showed that the generated VIC images significantly increase the number of extracted SIFT feature points. Furthermore, we have found that input color channel selection for the style translation affects the feature matching performance of the VIC images. Based on the investigation, we have found that translating from no-IC green-channel images to IC-sprayed red-channel images gives significant improvements to the SfM reconstruction quality. We have experimentally demonstrated that our new approach can reconstruct the whole stomach shapes of all seven subjects and showed that the estimated camera poses can be used for the lesion localization purpose. The reconstruction result videos can be accessed from the following link (\textcolor{blue}{http://www.ok.sc.e.titech.ac.jp/res/Stomach3D/}). In future work, we plan to investigate on how to increase the temporal pattern consistency of the VIC images to further improve the feature matching performance. 

\addtolength{\textheight}{-12cm}   









\bibliographystyle{IEEEtran}
\bibliography{main}

\end{document}